\documentclass[runningheads]{llncs}
\usepackage{graphicx}
\usepackage{tabularx}
\usepackage{tcolorbox}

\usepackage{tikz}
\usepackage{calc}
\usetikzlibrary{arrows.meta}
\usetikzlibrary{fit}
\usetikzlibrary{calc}
\usepackage{makecell}
\usepackage{colortbl}
\definecolor{mycolor}{rgb}{0.85, 0.95, 0.95}

\usepackage{float} 
\usepackage{tabularx}

\usepackage{amsmath, amsfonts, amssymb}
\usepackage{verbatim}
\usepackage{tabularx, booktabs}
\usepackage{graphicx}
\usepackage{tikz}
\usepackage{listings}
\usetikzlibrary{fit} 
 \usepackage{arydshln}
\newcommand{\squotes}[1]{`#1'}
\newcommand{\dquotes}[1]{``#1''}
\usepackage{graphicx}
\usepackage{subcaption}
\usepackage{makecell}

\usepackage{multirow}
\usepackage{ragged2e}  
\usepackage{booktabs}  
\usepackage{array, booktabs}
\usepackage{tabularx}
\usepackage{comment}
\usepackage{adjustbox}

\usepackage[utf8]{inputenc} 
\usepackage[T1]{fontenc}    
\usepackage{hyperref}       
\usepackage{url}            
\usepackage{amsfonts}       
\usepackage{nicefrac}       
\usepackage{microtype}      
\usepackage{xcolor}         
\usepackage{amsmath}
\usepackage{colortbl}
\definecolor{Gray}{gray}{1}
\usepackage{xcolor}
\definecolor{shadecolor}{rgb}{0.9,0.9,0.9}

\usepackage[framemethod=TikZ]{mdframed}
\usepackage{lipsum}

\mdfdefinestyle{PromptStyle}{
  linecolor=blue,
  outerlinewidth=2pt,
  roundcorner=20pt,
  innertopmargin=\baselineskip,
  innerbottommargin=\baselineskip,
  innerrightmargin=20pt,
  innerleftmargin=20pt,
  backgroundcolor=gray!50!white}
\newcommand{\RandomForest}{\texttt{RF~}}
\newcommand{\LogisticRegression}{\texttt{LR~}}
\newcommand{\MLPClassifier}{\texttt{MLP~}}
\newcommand{\KNeighbors}{\texttt{KNN~}}
\newcommand{\XGBoost}{\texttt{XGB~}}
\newcommand{\AdaBoost}{\texttt{AdaBoost~}}

\definecolor{bluegreen}{rgb}{0,0,0} 

\begin{document}
\title{\texttt{ChatGPT-HealthPrompt.} Harnessing the Power of XAI in Prompt-Based Healthcare Decision Support using ChatGPT}
\titlerunning{Harnessing the Power of XAI in Prompt-Based Healthcare Decision Support}
%

\renewcommand{\thefootnote}{}
\footnotetext{Accepted to the 3rd International workshop on \textit{Explainable and Interpretable Machine Learning (XI-ML@ECAI'23)}, co-located with ECAI 2023.}
\renewcommand{\thefootnote}{\arabic{footnote}}

\author{Fatemeh Nazary \and Yashar Deldjoo\thanks{Corresponding author.} \and
Tommaso Di Noia} 
\authorrunning{F. Nazary, Y.Deldjoo, T. Di Noia}
%
\institute{Polytechnic University of Bari, Italy \email{\{fatemeh.nazary,yashar.deldjoo,tommaso.dinoia\}@poliba.it}}

%
\maketitle              
\begin{abstract}
This study presents an innovative approach to the application of large language models (LLMs) in clinical decision-making, focusing on OpenAI's ChatGPT. Our approach introduces the use of \textit{contextual prompts}—strategically designed to include task description, feature description, and crucially, integration of domain knowledge—for high-quality binary classification tasks even in data-scarce scenarios. The novelty of our work lies in the utilization of domain knowledge, obtained from high-performing interpretable ML models, and its seamless incorporation into prompt design. By viewing these ML models as medical experts, we extract key insights on feature importance to aid in decision-making processes. This interplay of domain knowledge and AI holds significant promise in creating a more insightful diagnostic tool.
\vspace{2mm}

Additionally, our research explores the dynamics of zero-shot and few-shot prompt learning based on LLMs. By comparing the performance of OpenAI's ChatGPT with traditional supervised ML models in different data conditions, we aim to provide insights into the effectiveness of prompt engineering strategies under varied data availability. In essence, this paper bridges the gap between AI and healthcare, proposing a novel methodology for LLMs application in clinical decision support systems. It highlights the transformative potential of effective prompt design, domain knowledge integration, and flexible learning approaches in enhancing automated decision-making.

\end{abstract}

\section{Introduction} 
\textbf{Motivation.} The ever-evolving field of Natural Language Processing (NLP) has opened the door for potential advancements in a variety of sectors, the medical and healthcare field being no exception. The latest breakthroughs achieved by large language models (LLMs) such as OpenAI's GPT~\cite{brown2020language}, Google's PALM~\cite{chowdhery2022palm}, and Facebook's LaMDA~\cite{DBLP:journals/corr/abs-2201-08239}, has sparked intriguing speculation about the integration of AI in clinical decision-making and healthcare analytic. Consider a scenario where a healthcare professional, seeking a second opinion on a complex case, turns to an AI-powered system such as ChatGPT instead of consulting another colleague.  With the provision of all relevant medical data and context, the model could provide a comprehensive interpretation of the information, potentially suggesting diagnoses or treatment options. This application is no longer purely speculative; models such as OpenAI's ChatGPT have already demonstrated their potential to understand and generate contextually relevant responses, indicating a potential to become supportive aids in clinical decision-making.

Notwithstanding their great promise, it is important to underline that LLMs gain their power from being trained on billions of documents on internet data, allowing them to identify connections between words in various settings and formulate the most likely word sequences in a given new context. As promising as this might seem, the application of LLMs in the healthcare field is not \textit{without risks}, including potential inaccuracies due to a lack of specific medical training, misinterpretation of context, or data privacy concerns, all of which could have serious consequences in this critical domain. Enhancing the performance of LLMs for specific applications typically involves two major strategies, namely \textit{fine-tuning} and \textit{prompt design}~\cite{wang2023prompt,liu2023pre,white2023prompt}. Both serve similar goals in LLM enhancement to accomplish desired tasks, however, they differ significantly in their approaches. Prompting manipulates the model at inference time by providing context, instruction, and examples within the prompt, leaving the model's parameters unchanged. In contrast, fine-tuning modifies the model parameters according to a representative dataset, demanding more resources, however resulting in more specialized and consistent outcomes across similar tasks. 

Prompt-design strategies for LLMs can be categorized based on task complexity and the degrees of contextual examples provided. These categories include \textit{zero-shot}, \textit{one-shot}, and \textit{few-shot prompting}. Zero-shot prompting is ideal for straightforward, well-defined tasks that do not require multiple examples. It involves providing a single, concise prompt and relying on the model's pre-existing knowledge to generate responses. For instance, in the medical field, zero-shot prompting could be employed to provide a broad overview of common diseases. On the other hand, the techniques of one-shot and few-shot prompting involve guiding the model with one or more examples or queries to steer it toward generating desired outputs. An example prompt might be, \dquotes{\textit{Consider a 57-year-old male with high cholesterol, abnormal ECG, and exercise-induced angina, who shows signs of heart disease. Conversely, a 48-year-old male experiencing typical angina, but maintaining normal blood sugar levels and ECG, and without exercise-induced angina, is likely not suffering from heart disease. Based on these examples, predict the presence or absence of heart disease for a newly presented individual with specified medical conditions, using the narratives provided as guidance.}} Historically, GPT-$1$ was evaluated for its zero-shot capabilities, demonstrating encouraging results. As language models evolved, however, there was a shift towards the use of few-shot prompting in subsequent iterations such as GPT-$2$ and GPT-$3$ \cite{agrawal2022large}. Despite the success of these models, the format of the prompt and the sequence of examples can have a substantial impact on task performance \cite{bhatti2023art}. As a result, optimizing the use of prompts in these models continues to be an area of active research. While tasks requiring specialized skills often benefit from fine-tuning, this methodology is beyond the scope of our current research. Instead, our focus lies in employing \textbf{guided prompt-design} to improve decision-making processes within the medical domain.

\vspace{1mm}

\noindent \textbf{Contributions.} This paper aims to explore the application of OpenAI's ChatGPT to tackle binary classification tasks within clinical decision support systems, utilizing contextual prompts for high-quality predictions with minimal data. Our method underscores the incorporation of \dquotes{\textbf{domain-specific knowledge,}} extracted from interpretable ML models, to enhance prediction tasks and foster few-shot (and also zero-shot) learning. We showcase how OpenAI models can handle downstream tasks, matching the performance of traditional supervised ML models with ample data, even in data-scarce scenarios. We further discuss the relative advantages of zero-shot and few-shot prompts engineering.

Figure \ref{fig:model} symbolically presents our novel approach to enhancing medical decision-making by leveraging interpretable ML. The core novelty of our work lies in crafting effective prompts that will function as inputs for OpenAI's ChatGPT. To generate these prompts, we start with a basic version containing the task description. To provide more context, we integrate a feature description, which highlights key features relevant to the classification task at hand. Crucially, we also incorporate domain knowledge, obtained by separately training ML models and using their feature-based explanations as a source of expert insights. These ML models can be metaphorically seen as doctors, each emphasizing specific features deemed important for diagnosing particular diseases. This integration of expert-driven knowledge aims to enrich the diagnostic process. Further, we study the dynamics of zero-shot versus few-shot prompt engineering by varying the number of examples supplied to the prompt, thus enabling us to evaluate the system's adaptability to different data volumes. In summary, our contributions are summarized as follows:


\vspace{1mm}
\begin{figure}[t]
    \centering
    \includegraphics[trim=0 15pt 0 0pt, clip, width=0.98\linewidth]{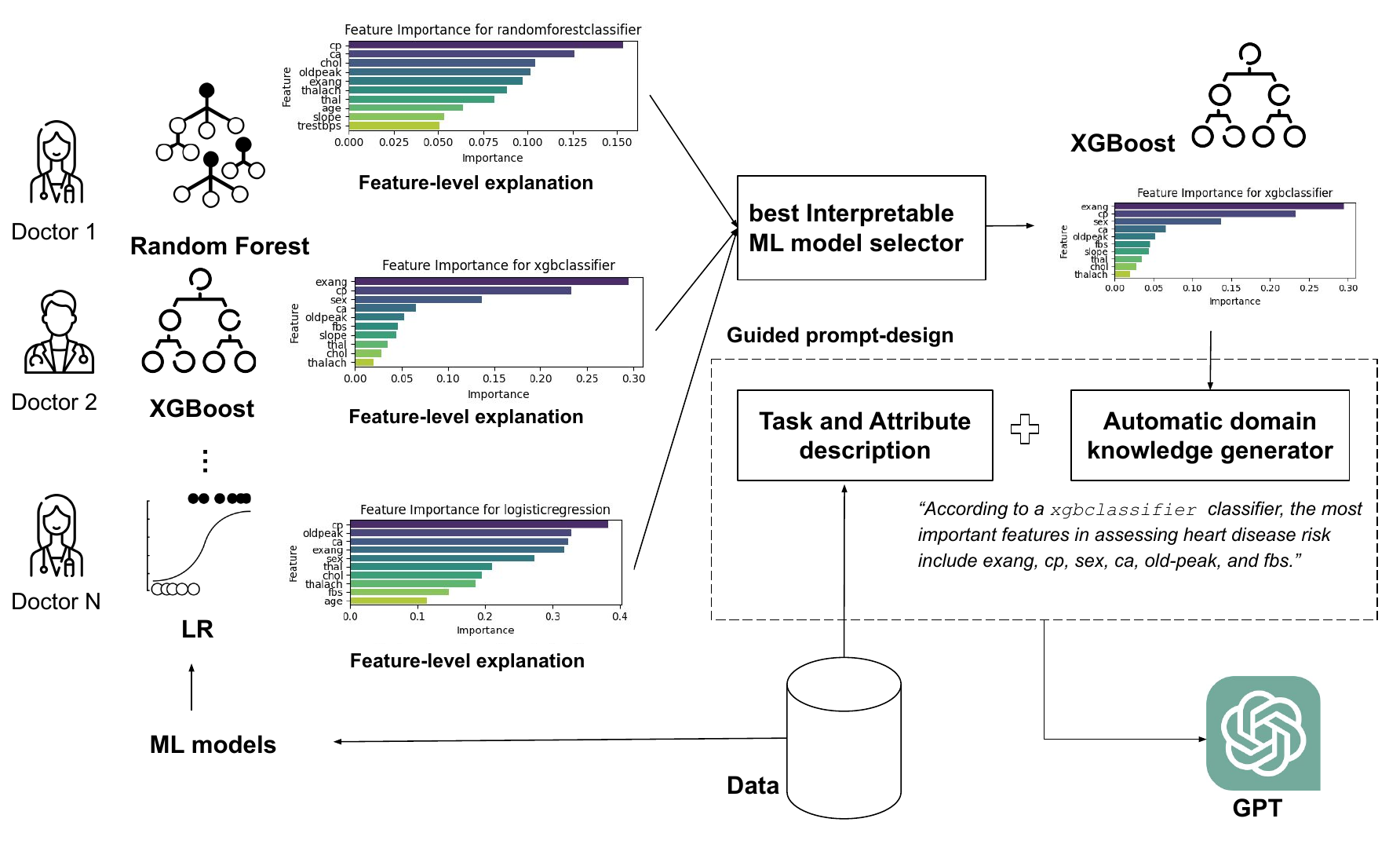}
\caption{Flowchart illustrating the conceptual framework of the paper}
\label{fig:model}
\end{figure}


\begin{itemize}
\item \textbf{Utilizing OpenAI's ChatGPT for Clinical Decision Support.} We exploit the potential of OpenAI's ChatGPT in clinical decision support systems, specifically for binary classification tasks, demonstrating its practical application in this domain.  \vspace{0.3mm}
\item \textbf{Advancing Prompt Engineering and Domain Knowledge Integration.} Our study introduces a novel approach to prompt engineering using \dquotes{contextual prompts} and underscores the integration of domain-specific knowledge. These carefully crafted prompts lead to high-quality predictions even in scenarios with limited data. We further enhance this process by creating \textbf{a domain knowledge generator}, which leverages high-performing ML tasks. We treat these ML models as metaphorical medical experts, enhancing prediction tasks and facilitating these models to operate as few-shot learners.
\item \textbf{Exploring Few-Shot vs. Zero-Shot Learning:} Our work contrasts the few-shot learning capability of OpenAI's ChatGPT with traditional supervised ML models, trained with ample data. We highlight the benefits of zero-shot and few-shot prompt engineering, shedding light on the interplay between data availability and prediction quality.

\end{itemize}

The structure of this paper is as follows: Section \ref{sec:rw} presents the related work, detailing the history of Transformers and Large Language Models, prompt engineering, and the use of LLMs in clinical decision-making. Section \ref{sec:openAIML} introduces our novel OpenAI-ML framework. Section \ref{sec:exp} outlines our experiments and methodology, followed by Section \ref{sec:result}, which discusses the performance outcomes and risks associated with our proposed system.

\section{Related Work}
\label{sec:rw}

\subsection{History of Transformers and Large Language Models}
The evolution of language models has been marked by the ongoing pursuit of more complex, versatile, and human-like machine representations of language. Prior to 2017, Natural Language Processing (NLP) models were largely trained on supervised learning tasks, limiting their generalizability \cite{liu2023pre}. However, the advent of the transformer architecture by Vaswani et al.~\cite{vaswani2017attention}, a self-attention network, led to the development of ground-breaking models such as Generative Pretrained Transformers (GPT) and Bidirectional Encoder Representations from Transformers (BERT) \cite{devlin2018bert}. These models use a semi-supervised approach, combining unsupervised pre-training with supervised fine-tuning, to achieve superior generalization capabilities. In recent years, we have witnessed a rapid progression in GPT models, resulting in the creation of GPT-3, a behemoth model with 175 billion parameters. Notwithstanding, these models still face significant challenges, including alignment with human values and the potential for generating biased or incorrect information. Efforts have been made to mitigate these issues, with the introduction of reinforcement learning from human feedback (RLHF) for improved model fine-tuning and alignment, as exemplified in the evolution of GPT-3 into ChatGPT \cite{liu2023deid}.

\subsection{Prompt Engineering}
Prompts play a crucial role in controlling and guiding the application of Large Language Models (LLMs). Essentially, a prompt is a set of instructions given to the model using natural, human language to define the task to be performed and the desired output. Prompts can be broadly categorized into two main types: \textit{manual prompts} and \textit{automated prompts} \cite{wang2023prompt}. Manual prompts are carefully designed by human specialists to provide models with precise instructions. However, their creation requires substantial expertise and time, and even minor adjustments can significantly affect the model's predictions. To overcome these limitations, various automated methods for prompt design have been developed.

Automated prompts, including discrete and continuous prompts, have gained popularity due to their efficiency and adaptability. They are generated using a variety of algorithms and techniques, thereby reducing the need for human intervention. Continuous prompts consider the current conversation context to generate accurate responses, while discrete prompts depend on predefined categories to produce responses. There are also both static and dynamic prompts that interpret the historical context differently. Ultimately, the performance and effectiveness of LLMs are significantly influenced by the quality and efficiency of these prompts \cite{liu2023pre}.

\subsection{Use of LLMs in Clinical Decision Making}
Prompt engineering and LLMs such as ChatGPT and GPT-4 have shown promising performance in advancing the medical field. Their diverse applications in various tasks, including unique prompt designs, multi-modal data processing, and deep reinforcement learning, are discussed comprehensively in Wang et al. \cite{wang2023prompt}. The LLMs demonstrate promising potential for clinical decision-making due to their adaptive abilities, enabling zero-shot and few-shot in-context learning despite the scarcity of labeled data. They contribute to offering diagnostic insights, treatment suggestions, and risk assessments. However, while these advancements demonstrate the transformative potential of LLMs in healthcare, it underlines the need for additional research to address prompt engineering limitations and ensure the ethical, reliable, safe, and effective use of LLMs in healthcare settings.

\vspace{-3mm}

\section{OpenAI-ML Framework for Health Risk Assessment}
\label{sec:openAIML}

\begin{figure}[!t]
\centering
\begin{tikzpicture}[
    scale = 0.62, 
    node distance = 2.55cm, 
    auto, 
    block/.style={rectangle, draw, text width=6em, text centered, rounded corners, minimum height=4em},
    line/.style={draw, -{Latex[length=2mm]}},
    short_line/.style={draw, -{Latex[length=2mm]}, shorten <=1cm, shorten >=0.21cm},
    dashed_block/.style={draw, dashed, inner sep=2em}
]
    \node [block, fill=blue!30] (input_data) {Input Data};
    \node [block, below of=input_data, fill=blue!30] (ml_training) {ML Model Training};
    \node [block, right of=ml_training, fill=red!30] (domain_generator) {Domain Knowledge Generator};
    \node [block, above of=domain_generator, fill=orange!30] (example_sampling) {Candidate Example Sampling};
    \node [block, above of=example_sampling, fill=orange!30] (task_description) {Task Description};
    \node [block, right of=task_description, fill=orange!30] (feature_specification) {Feature Specification};
    \node [block, below of=feature_specification, fill=yellow!30] (optimal_prompt) {Optimal Prompt};
    \node [block, right of=optimal_prompt, fill=green!30] (chatgpt) {ChatGPT};
    \node [block, right of=chatgpt, fill=green!30] (post_processing) {Post Processing};
    \node [block, right of=post_processing, fill=purple!30] (output) {Output};

    \path [line] (input_data) -- (ml_training);
    \path [line] (input_data.east) -| ($(example_sampling.west) - (0.5cm, 0)$) -- (example_sampling.west);
    \path [line] (ml_training) -- (domain_generator);
    \path [line] (task_description) -- (feature_specification);
    \path [line] (domain_generator) -- (optimal_prompt);
    \path [line] (feature_specification) -- (optimal_prompt);
    \path [line] (example_sampling.east) -- (optimal_prompt.west);
    \path [line] (optimal_prompt) -- (chatgpt);
    \path [line] (chatgpt) -- (post_processing);
    \path [line] (post_processing) -- (output);

    \node [dashed_block, fit=(domain_generator)(example_sampling)(task_description)(feature_specification)(optimal_prompt)] (box) {};
    \node [above=4.2cm] {{Guided Prompt Design}};
\end{tikzpicture}
\caption{Flowchart illustrating the proposed guided prompt design process for integrating contextual information in the medical field, emphasizing the sequential steps involved in designing effective prompts.}
\label{fig:updated_flowchart}
\end{figure}
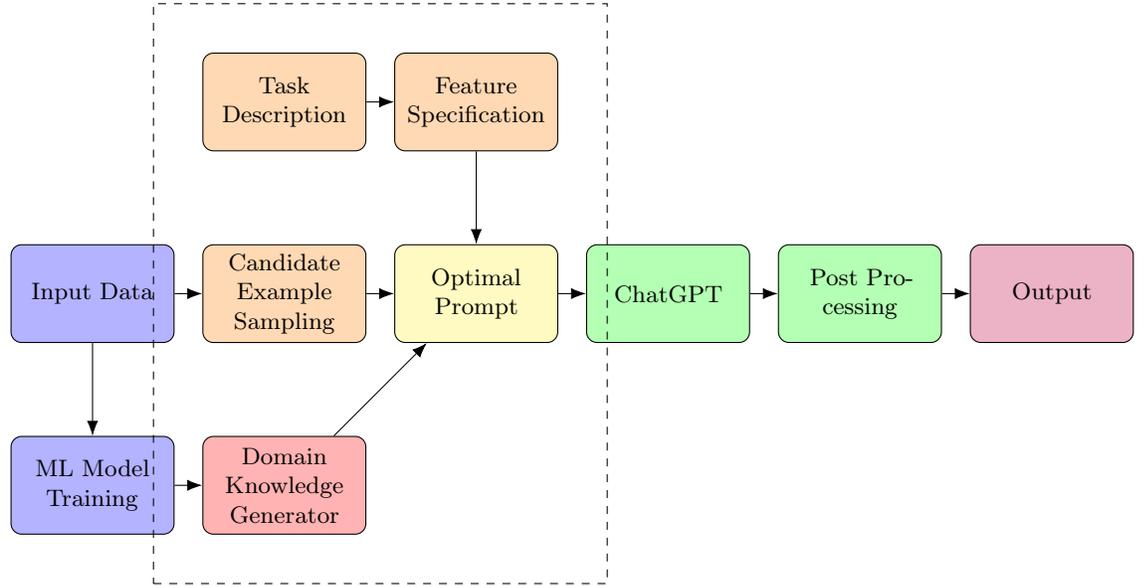

\noindent \textbf{Overview of the proposed system.} We propose a system for health risk assessment that leverages OpenAI's advanced language model, \texttt{ChatGPT-3.5-Turbo}. This system employs a conversation-based strategy to predict the risk of heart disease. It is designed to output binary responses (\squotes{1} or \squotes{0}), which correspond to high and low risk of heart disease, respectively. A detailed flowchart of our proposed system can be seen in Figure \ref{fig:updated_flowchart}.



\begin{itemize}
    \item \textbf{Part 1: Task Instruction:} The model is provided a task to assess heart disease risk based on given attributes. Here, the degree of diameter narrowing in the blood vessels informs the risk assessment, where less than 50\% narrowing indicates low risk (\squotes{0}) and more than 50\% indicates high risk (\squotes{1});
    
    \item \textbf{Part 2: Attribute Description:} The model is informed about the meaning of each attribute involved in the risk assessment. This ranges from the individual's age, sex, and chest pain type (cp), to their cholesterol levels (chol), among others. Each attribute is clearly defined, aiding the model to understand its relevance and role in the task at hand;

    \item \textbf{Part 3: In-context Examples:}  The model is given example scenarios with specific attribute inputs and corresponding risk assessment answers, aiding it to understand the pattern and relationship between the attributes and the risk level;

    \item \textbf{Part 4: Integration of Domain Knowledge:} The model is given domain knowledge which is simulated by best-performing interpretable ML models, such as \texttt{RandomForestClassifier}, \texttt{LogisticRegression}, and \texttt{XGBClassifier}. These models offer an ordered list of feature importance, which can aid the ChatGPT model in making a more informed assessment;

    \item \textbf{Part 5: Formulation of a Question/Problem:} The model is presented with an test instance. The instance involves specific inputs for each attribute and the model is asked to assess the risk level based on the prior instructions, examples, attribute descriptions, and domain knowledge. This allows for the practical application of the instruction in a real-world example.

\end{itemize}

\begin{table}[!t]
\caption{Summary of Domain Knowledge Types}
\centering
\begin{tabularx}{\textwidth}{>{\raggedright\arraybackslash}p{2.8cm} >{\raggedright\arraybackslash}X >{\raggedright\arraybackslash}X >{\raggedright\arraybackslash}X}
\toprule
& \textbf{dk0} & \textbf{Odd dk (MLFI)} & \textbf{Even dk (MLFI-ord)} \\
\midrule
\textbf{Name} & N/A & ML defines feature importance & Similar to MLFI, includes feature order \\
\addlinespace
\textbf{Description} & No extra domain knowledge & Feature importance defined by ML algorithms & Includes both feature importance and order \\
\addlinespace
\textbf{Focus} & N/A & Feature Attribution & Feature Attribution with order-awareness \\
\addlinespace
\textbf{Implementation} & Solely data-driven & XGB, RF, Ada, LR & XGB, RF, Ada, LR \\
\addlinespace
\textbf{Usage Scenarios} & \textit{Zero-shot} scenarios; Simple tasks with no specific domain knowledge required & \textit{Few-shot} scenarios; Tasks requiring the understanding of feature importance & \textit{Few-shot} scenarios; Tasks requiring the understanding of both feature importance and order \\
\bottomrule
\end{tabularx}
\label{tab:domain_kw}
\end{table}

\noindent \textbf{Domain Knowledge Integration.}
Domain knowledge integration is categorized into three types as shown in Table \ref{tab:domain_kw}. These types are designed to handle different tasks based on the requirements of domain knowledge and the order of features. 

\begin{itemize}
    \item \textbf{dk0:} This scenario involves no additional domain knowledge, where learning is purely data-driven. It is most suitable for simple tasks that do not require specific domain knowledge, making it an effective choice for testing \textit{zero-shot} scenarios; \\ \vspace{-2mm}
    
    \item \textbf{Odd dk or MLFI:} This category identifies the importance of features as determined by various ML algorithms such as XGB, RF, Ada, and LR. It is particularly useful in tasks requiring the understanding of feature importance, and it is used to test \textit{few-shot} scenarios (MLFI is short for Machine learning Feature Importance); \\ \vspace{-2mm}
    
    \item \textbf{Even dk or MLFI-ord:} This category not only considers the importance of features but also their order, as determined by ML algorithms. It is ideal for tasks that require an understanding of both feature importance and their order, and it is also used to test \textit{few-shot} scenarios (MLFI is short for Machine learning Feature Importance-\textit{ordered}). \vspace{-2mm}
\end{itemize}

\noindent More detailed information regarding the conversation used is depicted below.

\usetikzlibrary{shapes.geometric, arrows.meta, fit, calc}

\mdfdefinestyle{PartOneStyle}{
  linecolor=blue,
  outerlinewidth=2pt,
  roundcorner=20pt,
  innertopmargin=10pt,
  innerbottommargin=10pt,
  innerrightmargin=10pt,
  innerleftmargin=10pt,
  backgroundcolor=green!20!white,
  frametitle={\textbf{Part 1: Task Instruction}}
}

\mdfdefinestyle{PartTwoStyle}{
  linecolor=blue!60!white,
  outerlinewidth=2pt,
  roundcorner=20pt,
  innertopmargin=10pt,
  innerbottommargin=10pt,
  innerrightmargin=10pt,
  innerleftmargin=10pt,
  backgroundcolor=red!20!white,
  frametitle={\textbf{Part 3: In-Context Example}}
}
\mdfdefinestyle{PartThreeStyle}{
  linecolor=blue!60!white,
  outerlinewidth=2pt,
  roundcorner=20pt,
  innertopmargin=10pt,
  innerbottommargin=10pt,
  innerrightmargin=10pt,
  innerleftmargin=10pt,
  backgroundcolor=yellow!20!white,
frametitle={\textbf{Part 2: Attribute Description}}
}
\mdfdefinestyle{PartFourStyle}{
  linecolor=blue!60!white,
  outerlinewidth=2pt,
  roundcorner=20pt,
  innertopmargin=10pt,
  innerbottommargin=10pt,
  innerrightmargin=10pt,
  innerleftmargin=10pt,
  backgroundcolor=orange!20!white,
  frametitle={\textbf{Part 4: Domain Knowledge Integration using Interpretable ML}}
}
\mdfdefinestyle{PartFiveStyle}{
  linecolor=blue,
  outerlinewidth=2pt,
  roundcorner=20pt,
  innertopmargin=10pt,
  innerbottommargin=10pt,
  innerrightmargin=10pt,
  innerleftmargin=10pt,
  backgroundcolor=purple!20!white,
  frametitle={\textbf{Part 5: Final Task Question}}
}

\begin{mdframed}[style=PartOneStyle]
Given the provided input attributes, evaluate the risk of heart disease for the individual.
The diagnosis of heart disease (angiographic disease status) is based on the degree of diameter narrowing in the blood vessels:
\begin{itemize}
    \item 0: Less than 50\% diameter narrowing, implying a lower risk.
    \item 1: More than 50\% diameter narrowing, indicating a higher risk.
\end{itemize}
If the assessment determines a high risk, the output should be \squotes{1}. If the risk is determined to be low, the output should be \squotes{0}.
Evaluate the credit risk based on given attributes. If good, respond with \squotes{1}, if bad, respond with \squotes{0}.
\end{mdframed}

\begin{mdframed}[style=PartThreeStyle]
The explanation of each attribute is as follows:
\begin{itemize}
    \item Age: Age of the individual
    \item Sex: Sex of the individual (1 = Male, 0 = Female)
  \item Cp: Chest pain type (1 = typical angina, 2 = atypical angina, 3 = non-anginal pain, 4 = asymptomatic)
  \item Trestbps: Resting blood pressure (in mm Hg on admission to the hospital)
  \item Chol: Serum cholesterol in mg/dl
  \item Fbs: Fasting blood sugar > 120 mg/dl (1 = true, 0 = false)
  \item Restecg: Resting electrocardiographic results (0 = normal, 1 = having ST-T wave abnormality, 2 = showing probable or definite left ventricular hypertrophy)
  \item Thalach: Maximum heart rate achieved
  \item Exang: Exercise-induced angina (1 = yes, 0 = no)
  \item Oldpeak: ST depression induced by exercise relative to rest
  \item Slope: The slope of the peak exercise ST segment (1 = upsloping, 2 = flat, 3 = downsloping)
  \item Ca: Number of major vessels (0-3) colored by fluoroscopy
  \item Thal: Thalassemia (3 = normal, 6 = fixed defect, 7 = reversible defect)
\end{itemize}
\end{mdframed}
\vspace{-4mm}
\begin{mdframed}[style=PartTwoStyle]
Example 1: \\
<Inputs 1>: age: 57, sex: 1, cp: 2, trestbps: 140, chol: 265, fbs: 0, restecg: 1, thalach: 145, exang: 1, oldpeak: 1, slope: 2, ca: 0.2, thal: 5.8 \\
<Answer 1>: 1 \\ \vspace{-4mm}

\noindent Example 2: \\
<Inputs 2>: age: 48, sex: 1, cp: 2, trestbps: 130, chol: 245, fbs: 0, restecg: 0, thalach: 160, exang: 0, oldpeak: 0, slope: 1.4, ca: 0.2, thal: 4.6 \\
<Answer 2>: 0 \\ \vspace{-4mm}

\noindent Example 3: \\
<Inputs 3>: age: 44, sex: 1, cp: 4, trestbps: 112, chol: 290, fbs: 0, restecg: 2, thalach: 153, exang: 0, oldpeak: 0, slope: 1, ca: 1, thal: 3 \\
<Answer 3>: 1 \\ \vspace{-4mm}

\end{mdframed}
\vspace{-4mm}

\begin{mdframed}[style=PartFourStyle]
\vspace{-3mm}
Domain Knowledge: \vspace{-3mm}
\begin{itemize}
    \item \textbf{dk0}: None \\
    \vspace{-4mm}
\item \textbf{dk1}: According to a \texttt{randomforestclassifier} classifier, the most important features in assessing heart disease risk include cp, ca, chol, oldpeak, exang, and thalach. Features like fbs and restecg have relatively lower importance;  \\ \vspace{-3mm}
\item \textbf{dk2}: The \textbf{order of features} is critically important when evaluating heart disease risk. The sequence of features according to their importance starts with cp, followed by ca, then chol, oldpeak, exang, thalach, thal, age, slope, trestbps, sex, fbs, and finally restecg; \\
    \vspace{-3mm}
\item \textbf{dk3}: According to a \texttt{logisticregression} classifier, the most important features in assessing heart disease risk include cp, oldpeak, ca, exang, sex, and thal. Features like restecg and trestbps have relatively lower importance; \\
    \vspace{-3mm}
\item \textbf{dk4}: The \textbf{order of features} is critically important when evaluating heart disease risk. The sequence of features according to their importance starts with cp, followed by oldpeak, then ca, exang, sex, thal, chol, thalach, fbs, age, slope, restecg, and finally trestbps; \\
    \vspace{-3mm}
\item \textbf{dk5}: According to a \texttt{xgbclassifier} classifier, the most important features in assessing heart disease risk include exang, cp, sex, ca, oldpeak, and fbs. Features like trestbps and restecg have relatively lower importance; \\
    \vspace{-3mm}
\item \textbf{dk6}: The \textbf{order of features} is critically important when evaluating heart disease risk. The sequence of features according to their importance starts with exang, followed by cp, then sex, ca, oldpeak, fbs, slope, thal, chol, thalach, age, trestbps, and finally restecg;
\end{itemize}

\end{mdframed}

\begin{mdframed}[style=PartFiveStyle]
\vspace{-1mm}
 Now, given the following inputs, please evaluate the risk of heart disease: \\
\noindent <Inputs>: age: 46.0, sex: 1.0, cp: 3.0, trestbps: 150.0, chol: 163.0, fbs: 0.2, restecg: 0.0, thalach: 116.0, exang: 0.0, oldpeak: 0.0, slope: 2.2, ca: 0.4, thal: 6.2 \\
\noindent  <Answer>: ?

\end{mdframed}

\begin{table}[!b]
\caption{Summary of ML models and their Hyper-parameters}
\centering
\begin{tabular}{|l|p{0.7\linewidth}|c|}
\hline
\textbf{Algorithm} & \textbf{Parameters} & \textbf{Total} \\
\hline
RF & n\_estimators, max\_depth, min\_samples\_split, min\_samples\_leaf, bootstrap & 5 \\
\hline
LR & C, penalty, solver & 3 \\
\hline
MLP & hidden\_layer\_sizes, activation, solver, alpha, learning\_rate, learning\_rate\_init, tol, max\_iter & 8 \\
\hline
KNeighbors & n\_neighbors, weights, algorithm, leaf\_size, p & 5 \\
\hline
XGB & use\_label\_encoder, eval\_metric, n\_estimators, learning\_rate, max\_depth, colsample\_bytree & 6 \\
\hline
AdaBoost & n\_estimators, learning\_rate & 2 \\
\hline
\end{tabular}
\label{tab:algo_parameters}
\end{table}

\section{Experimental Setup}
\label{sec:exp}

\noindent \textbf{Task.} This work focuses on binary classification within a health risk assessment context in machine learning (ML). The task involves learning a function $f: \mathcal{X} \rightarrow \{0,1\}$, predicting a binary outcome $y \in \{0,1\}$ for each feature instance $x \in \mathcal{X}$.

\vspace{1.5mm}
\noindent \textbf{Dataset.} We used the Heart Disease dataset \cite{janosi1988heart} was collected from four hospitals located in the USA, Switzerland, and Hungary.  The task associated with this dataset is binary classification, aimed at determining the presence or absence of heart disease. The dataset initially contains 621 out of 920 samples with missing values, which were all successfully \textit{imputed} using the \texttt{KNN} method, resulting in no samples with missing values. The distribution of males to females in the dataset is approximately 78\% to 21\%, while the disease prevalence ratio between males and females is approximately 63\% to 28\% in the given population, males tend to suffer from heart disease more frequently than females).
\vspace{1.5mm}

\noindent \textbf{Baseline Prediction Models.} As a comparison, we also implemented three basic prediction models, namely \texttt{majority class-1} prediction, \texttt{majority class-0} prediction, and \texttt{random} prediction. These models are non-informed baselines and operate under a predefined mode and generate baseline predictions without relying on input features.

\vspace{0.3mm}
\noindent \textbf{Hyperparameters and ML Models.} We employed six ML models, each with a distinct set of hyperparameters. These were optimized using a randomized search cross-validation (CV) strategy, with a total of 29 unique hyperparameters across all models. This led to an extensive model tuning process involving numerous model iterations. We used a 5-fold CV (0.8, 0.2) with \textit{RandomizedSearchCV} over 20 iterations. The exact hyperparameters depended on the specific model and are listed in Table~\ref{tab:algo_parameters}. 

\vspace{0.3mm}
\noindent \textbf{OpenAI Model.} We utilized the OpenAI's \texttt{GPT-3.5-turbo} model to perform heart disease risk assessment based on patient health data. The model interacts through a chat-like interface, taking a sequence of messages and user inputs as prompts and generating corresponding output responses. The model was configured with a temperature setting of 0, dictating no randomness of the model's responses. The usage of the OpenAI model involved the generation of requests, comprising individual patient data features. Each request was processed by the model, returning an output prediction. For efficiency, these requests were sent in batches. The Python code made use of the OpenAI API, the tenacity library for retrying failed requests, and other common data processing libraries such as pandas. The code is available for reproducibility at this link.\footnote{Source code.\url{https://github.com/atenanaz/ChatGPT-HealthPrompt}}





\section{Results}
\label{sec:result}

Table \ref{tab:results_table} provides an in-depth comparison of heart health risk prediction performance using traditional ML models and OpenAI-based predictions. We have excluded results from \texttt{Random}, \texttt{Maj0}, and \texttt{Maj1} for average calculations and highlighted utilized values in the table for better tracking. We evaluated OpenAI predictions against conventional models such as \RandomForest, \LogisticRegression, \MLPClassifier, \KNeighbors, \XGBoost, \AdaBoost, with prompts categorized based on the integration of extra domain knowledge or without it. We also considered the number of examples used in prompt formulation (represented as $N_{ex}$). The results are presented in two main dimensions: \textit{overall performance} (F1 and Acc.) and \textit{risks} (FPC, FNC, and cost-sensitive accuracy). In this context, a particular emphasis was placed on false-negative costs due to the significant health risk of incorrectly diagnosing a healthy individual as sick. Lastly, we introduced a new metric, cost-sensitive accuracy, with specific weights (FP=0.2, FN=0.8) that may vary based on the scenario.

\vspace{-3mm}

\subsection{Overall performance}
The section sheds light on the comparative performance between classical machine learning (ML) models and OpenAI-based models, with and without the incorporation of domain knowledge as proposed in this paper (cf. Section \ref{sec:openAIML}).\vspace{1mm}

\noindent \textbf{Classical ML vs. OpenAI without domain knowledge.} From the provided table, it is clear that the OpenAI-based models using prompts have produced varied results depending on the number of examples used in their construction (denoted by $N_{ex}$). Specifically, looking at \texttt{prompt-0} which uses no extra domain knowledge, we see that its performance significantly improves as $N_{ex}$ increases. At \squotes{$N_{ex}=0$}, the F1 score is 0.7402 and the Accuracy is 0.6413, which are better than those of \texttt{Maj-1}, \texttt{Maj-0}, and \texttt{random}, indicating an advancement of OpenAI-ML in \textit{zero-shot scenarios }over those basic models. Yet, they are still below the average of classical Machine Learning models.

Progressing forward, in \textit{few-shot scenarios}, there is a noticeable improvement in the performance of \texttt{prompt-0}, which does not leverage any domain knowledge. As $N_{ex}$ (the number of examples) increases, this model even surpasses some traditional machine learning counterparts. At  \squotes{n=16}, it reaches an F1 score of 0.8241 and an Accuracy of 0.7935, which are closely comparable to the average F1 (0.8576) and Accuracy (0.8203) of classical ML models.  This showcases the robustness and potential of the OpenAI-based prediction model when a larger number of examples are used for prompt construction. It demonstrates that while classical machine learning methods have more consistent performance, the OpenAI models have the ability to learn and improve from more example prompts, achieving competitive performance with an increasing number of examples.

The study underscores the potential of prediction models based on OpenAI. Even though these models initially perform at a lower level compared to traditional ML models, the inherent iterative learning and enhancement capabilities of OpenAI models become increasingly clear as the number of examples used for prompt construction increases.

\vspace{2mm}
\noindent \textbf{Classical ML vs. OpenAI-ML \underline{with} domain knowledge.} The comparison between OpenAI's GPT-3.5 and classical machine learning models reveals intriguing insights. The OpenAI prediction prompts adopt two strategies: some without extra domain knowledge, while others integrate results from classical ML models (\RandomForest, \LogisticRegression, \XGBoost). We selected these models due to their capability to yield attribute-based explanation outcomes, as illustrated in figure \ref{fig:model}, and their inherent diversity. The sequence of features is contemplated in half of these prompts, particularly the even-numbered ones (\texttt{prompt-2}, \texttt{prompt-4}, and \texttt{prompt-6}). Moving forward to the analysis, in the second type of prompts (\texttt{prompt-1}, \texttt{prompt-3}, and \texttt{prompt-5}) that utilize the prediction results of classical ML models without considering their order of importance, there is a noticeable improvement as n increases, similar to \texttt{prompt-0}, however, the increase is more drastic. For instance, at \squotes{n=8}, \texttt{prompt-5} outperforms all models, yielding the highest F1 score (0.8711), Accuracy (0.8424), clearly surpassing \underline{all} the baseline ML models. Note that here, using just 8 examples (about 2-3\% of the total training data) seems to be enough for the model to significantly outperform the average results of classical ML models, particularly in terms of cost-sensitive accuracy.

Lastly, the prompts where the order of features was considered (\texttt{prompt-2}, \texttt{prompt-4}, and \texttt{prompt-6}) generally show a similar pattern, with performance improving as n increases. However, their results appear to be slightly lower than the prompts using classical ML model results without considering feature order. This may suggest that for certain tasks, the added complexity of considering feature order does not always translate into a clear performance advantage.


\begin{tcolorbox}[colback=green!5!white,colframe=green!75!black]
\textbf{Summary.} This study contrasted the performance of classical Machine Learning (ML) models and OpenAI-based models, with and without the integration of domain knowledge. The initial performance of OpenAI models is lower than classical ML models in \textit{zero-shot scenarios}, but exhibits substantial enhancement with an increase in the number of examples, i.e., in \textit{few-shot scenarios}, eventually attaining comparable metric values. Upon \textbf{domain knowledge} integration, particularly the prediction results of classical ML models, OpenAI models show \textbf{significant performance} improvement, with some surpassing all baseline ML models. However, the benefit of incorporating feature order is not always clear.
\end{tcolorbox}
\vspace{-4.5mm}
\subsection{Risks} It can be observed that on average, False Negatives (FN) in a majority of experimental cases of OpenAI remain below those of the classical Machine Learning (ML) models. For example, OpenAI models recorded FN of 3.08, 10.7, 8.6, and 9.37 compared to an average FN of 10.9 for ML models. This advantage, however, comes with a trade-off of higher False Positives (FP). OpenAI models on average scored higher in FP with values such as 12.77, 8.971, 80.578, and 6.4571, compared to the classical ML models.

Interestingly, the best-performing prompts within the OpenAI models demonstrated very low FP and FN rates (e.g., 4.6 and 4.8), which in terms of FN, remain much lower than even the best ML models. Summarizing the key observations, it can be stated that while OpenAI models may present better results in specific cases, care should be taken when discussing the risk of these models' predictions in clinical decision support. This caution is due to the high variability and variance these models show (e.g., in one case, FN reaches 20.8), and simply considering average statistics may not provide an accurate representation of their performance. On the other hand, ML models show more homogeneous performance. This phenomenon might be attributed to the tendency of prompt-based predictions to produce more 1s than 0s, thereby decreasing FN. However, considering the accuracies, it is evident that the results are not randomly generated, and OpenAI models are indeed capable of making sense of the data. This demonstrates the potential power of these models but also the need for careful design and implementation in clinical decision-making scenarios.
\vspace{-2mm}

\begin{tcolorbox}[colback=green!5!white,colframe=green!75!black]
\textbf{Summary.} 
The study of risk and false predictions shows that while OpenAI models on average produced fewer False Negatives (FN) compared to traditional Machine Learning (ML) models, they came with a significant trade-off of higher False Positives (FP). Notably, the OpenAI models demonstrated high variability in their results, indicating that relying on average statistics may not provide a comprehensive view of their performance. The observations underscore the need for careful design and implementation of OpenAI models in clinical decision-making scenarios, especially considering their potentially higher risk of incorrect predictions.
\end{tcolorbox}

\begin{table}[H]
\centering
\caption{Performance Comparison Between Classical and OpenAI ML Models.}
\label{tab:results_table}
\begin{tabular}{lccccccccccc}
\toprule
Model & \makecell{DK \\ Type}  & \makecell{DK \\ source} & \makecell{$N_{ex}$} & \textbf{Pre.}$\uparrow$ & \textbf{Rec}$\uparrow$ & \textbf{F1}$\uparrow$ & \textbf{Acc.}$\uparrow$ & \makecell{\textbf{FP}\\ \textbf{Cost}}$ \downarrow$ & \makecell{\textbf{FN}\\ \textbf{Cost}}$ \downarrow$ & \makecell{\textbf{Cost-Sens} \\\textbf{Acc.}}$\uparrow$  \\
\toprule
\texttt{RF} & & & & 0.8585 & 0.875 & 0.8667 & 0.8478 & 3.0 & 10.4 & 0.9208 \\
\texttt{LR} & & & & 0.8241 & 0.8558 & 0.8396 & 0.8152 & 3.8 & 12.0 & 0.9047 \\
\texttt{MLP} & & & & 0.8381 & 0.8462 & 0.8421 & 0.8207 & 3.4 & 12.8 & 0.9031 \\
\texttt{KNN} & & & & 0.8654 & 0.8654 & 0.8654 & 0.8478 & 2.8 & 11.2 & 0.9176 \\
\texttt{XGB} & & & & 0.8667 & 0.875 & \textbf{0.8708} & \textbf{\color{black}{0.8533}} & 2.8 & 10.4 & 0.9224 \\
\texttt{AdaBoost}  & & & & 0.8304 & 0.8942 & 0.8611 & 0.8370 & 3.8 & 8.8 & \textbf{0.9243} \\
\texttt{Maj1} & & & & 0.5576 & 0.8846 & \cellcolor{mycolor}0.6840 & \cellcolor{mycolor}0.5380 & 14.6 & 9.6 & 0.8035 \\
\texttt{Maj0} & & & & 0.6842 & 0.125 & \cellcolor{mycolor}0.2114 & \cellcolor{mycolor}0.4728 & 1.2 & 72.8 & 0.5403 \\
\texttt{random} & & & & 0.5326 & 0.4712 & \cellcolor{mycolor}0.5 & \cellcolor{mycolor}0.4674 & 8.6 & 44.0 & 0.6204 \\ \midrule
\texttt{Avg.} & & & & 0.8472 & 0.8686 & \cellcolor{mycolor}0.8576 & \cellcolor{mycolor}0.8368 & \cellcolor{mycolor}3.26 & \cellcolor{mycolor} 10.9 & 0.9155 \\\midrule
\texttt{prompt-0} & \scriptsize{NO} & \texttt{-} &0 & 0.6267 & 0.9038 & \cellcolor{mycolor}0.7402 & \cellcolor{mycolor}0.6413 & 11.2 & 8.0 & 0.8600 \\
\texttt{prompt-1} & \scriptsize{MLFI} & \texttt{RF} &0 & 0.6121 & 0.9712 & 0.7509 & 0.6359 & 12.8 & 2.4 & 0.8850 \\
\texttt{prompt-2} & \scriptsize{MLFI-ord} & \texttt{RF} &0 & 0.5976 & 0.9712 & 0.7399 & 0.6141 & 13.6 & 2.4 & 0.8759 \\
\texttt{prompt-3} & \scriptsize{MLFI} & \texttt{LR}  &0 & 0.6242 & 0.9904 & 0.7658 & 0.6576 & 12.4 & 0.8 & 0.9016 \\
\texttt{prompt-4} & \scriptsize{MLFI-ord} & \texttt{LR} &0 & 0.6111 & 0.9520 & 0.7444 & 0.6304 & 12.6 & 4.0 & 0.8748 \\
\texttt{prompt-5} & \scriptsize{MLFI} & \texttt{XGB} &0 & 0.6108 & 0.9808 & 0.7528 & 0.6359 & 13.0 & 1.6 & 0.8890 \\
\texttt{prompt-6}& \scriptsize{MLFI-ord} & \texttt{XGB}&0 & 0.5941 & 0.9712 & 0.7372 & 0.6087 & 13.8 & 2.4 & 0.8736 \\ \midrule
\texttt{Avg.} & - & - & 2 & 0.6109 & 0.9629 & 0.7473 & 0.6319 & \cellcolor{mycolor}12.77 & \cellcolor{mycolor}3.08 & 0.8799 \\ \midrule

\texttt{prompt-0} & \scriptsize{NO} & \texttt{-}  &2  & 0.6375 & 0.9808 & 0.7727 & 0.6739 & 11.6 & 1.6 & 0.9037 \\
\texttt{prompt-1} & \scriptsize{MLFI} & \texttt{RF}  &2  & 0.6415 & 0.9808 & 0.7757 & 0.6793 & 11.4 & 1.6 & 0.9057 \\
\texttt{prompt-2} & \scriptsize{MLFI-ord} & \texttt{RF}  &2  & 0.6358 & 0.9904 & 0.7744 & 0.6739 & 11.8 & 0.8 & 0.9077 \\
\texttt{prompt-3} & \scriptsize{MLFI} & \texttt{LR}  &2  & 0.6159 & 0.9712 & 0.7537 & 0.6413 & 12.6 & 2.4 & 0.8872 \\
\texttt{prompt-4} & \scriptsize{MLFI-ord} & \texttt{LR}  &2  & 0.6711 & 0.9615 & 0.7905 & 0.7120 & 9.8 & 3.2 & 0.9097 \\
\texttt{prompt-5} & \scriptsize{MLFI} & \texttt{XGB}  &2  & 0.8548 & 0.5096 & 0.6386 & 0.6739 & 1.8 & 40.8 & 0.7442 \\
\texttt{prompt-6} & \scriptsize{MLFI-ord} & \texttt{XGB} &2  & 0.7935 & 0.7019 & 0.7449 & 0.7283 & 3.8 & 24.8 & 0.8241 \\ \midrule
\texttt{Avg.} & - & - & 2 & 0.6928 & 0.8708 & 0.7500 & 0.6832 & \cellcolor{mycolor}8.971 & \cellcolor{mycolor} 10.74 & 0.8689 \\ \midrule
\texttt{prompt-0} & \scriptsize{NO} & \texttt{-}   &4  & 0.6978 & 0.9327 & 0.7984 & 0.7337 & 8.4 & 5.6 & 0.9060 \\
\texttt{prompt-1} & \scriptsize{MLFI} & \texttt{RF}  &4  & 0.8659 & 0.6827 & 0.7634 & 0.7609 & 2.2 & 26.4 & 0.8303 \\
\texttt{prompt-2} & \scriptsize{MLFI-ord} & \texttt{RF}  &4  & 0.5886 & 0.9904 & 0.7384 & 0.6033 & 14.4 & 0.8 & 0.8795 \\
\texttt{prompt-3} & \scriptsize{MLFI} & \texttt{LR}  &4  & 0.8257 & 0.8654 & 0.8451 & 0.8207 & 3.8 &11.2 & 0.9096 \\
\texttt{prompt-4} & \scriptsize{MLFI-ord} & \texttt{LR}  &4  & 0.6375 & 0.9808 & 0.7727 & 0.6739 & 11.6 & 1.6 & 0.9037 \\
\texttt{prompt-5} & \scriptsize{MLFI} & \texttt{XGB}  &4  & 0.6205 & 0.9904 & 0.7630 & 0.6522 & 12.6 & 0.8 & 0.8995 \\
\texttt{prompt-6}& \scriptsize{MLFI-ord} & \texttt{XGB}  &4  & 0.8350 & 0.8269 & 0.8309 & 0.8098 & 3.4 & 14.4 & 0.8932 \\ \midrule
\texttt{Avg.} & - & - & 4 & 0.7244 & 0.8956 & 0.7874 & 0.7220 & \cellcolor{mycolor}8.058 & \cellcolor{mycolor} 8.6857 & 0.8888 \\
\midrule
\texttt{prompt-0} & \scriptsize{NO} & \texttt{-}  & 8  & 0.8041 & 0.7500 & 0.7761 & 0.7554 & 3.8000 & \cellcolor{mycolor}20.8000 & 0.8496 \\
\texttt{prompt-1}& \scriptsize{MLFI} & \texttt{RF}  & 8  & 0.8515 & 0.8269 & 0.8390 & 0.8207 & 3.0000 & 14.4000 & 0.8967 \\
\texttt{prompt-2}& \scriptsize{MLFI-ord} & \texttt{RF}  & 8  & 0.6944 & 0.9615 & 0.8065 & 0.7391 & 8.8000 & 3.2000 & 0.9189 \\
\texttt{prompt-3} & \scriptsize{MLFI} & \texttt{LR}  & 8  & 0.6242 & 0.9904 & 0.7658 & 0.6576 & 12.4000 & 0.8000 & 0.9016 \\
\texttt{prompt-4} & \scriptsize{MLFI-ord} & \texttt{LR} & 8  & 0.6776 & 0.9904 & 0.8047 & 0.7283 & 9.8000 & 0.8000 & 0.9267 \\
\texttt{prompt-5} & \scriptsize{MLFI} & \texttt{XGB}  & 8  & 0.8099 & 0.9423 & \textit{\color{blue}{0.8711}} & \textit{\color{blue}{0.8424}} & \cellcolor{mycolor}4.6000 & \cellcolor{mycolor}4.8000 & \textit{\color{blue}{0.9428}} \\
\texttt{prompt-6} & \scriptsize{MLFI-ord} & \texttt{XGB} & 8  & 0.8478 & 0.7500 & 0.7959 & 0.7826 & 2.8000 & \cellcolor{mycolor}20.8000 & 0.8592 \\ \midrule
\texttt{Avg.} & - & - & 8  & 0.7585 & 0.8873 & 0.8084 & 0.7608 & \cellcolor{mycolor}6.4571 & \cellcolor{mycolor} 9.3714 & 0.8993 \\
\midrule
\texttt{prompt-0} & \scriptsize{NO} & \texttt{-}   & 16 & 0.7946 & 0.8558 & \cellcolor{mycolor}0.8241 & \cellcolor{mycolor}0.7935 & 4.6000 & 12.0000 & 0.8979 \\
\texttt{prompt-1} & \scriptsize{MLFI} & \texttt{RF} & 16 & 0.7965 & 0.8654 & 0.8295 & 0.7989 & 4.6000 & 11.2000 & 0.9029 \\
\texttt{prompt-2} & \scriptsize{MLFI-ord} & \texttt{RF}  & 16 & 0.7538 & 0.9423 & 0.8376 & 0.7935 & 6.4000 & 4.8000 & 0.9288 \\
\texttt{prompt-3} & \scriptsize{MLFI} & \texttt{LR} & 16 & 0.8554 & 0.6827 & 0.7594 & 0.7554 & 2.4000 & 26.4000 & 0.8284 \\
\texttt{prompt-4} & \scriptsize{MLFI-ord} & \texttt{LR} & 16 & 0.7339 & 0.8750 & 0.7982 & 0.7500 & 6.6000 & 10.4000 & 0.8903 \\
\texttt{prompt-5} & \scriptsize{MLFI} & \texttt{XGB}  & 16 & 0.7638 & 0.9327 & 0.8398 & 0.7989 & 6.0000 & 5.6000 & 0.9269 \\
\texttt{prompt-6} & \scriptsize{MLFI-ord} & \texttt{XGB}  & 16 & 0.8198 & 0.8750 & 0.8465 & 0.8207 & 4.0000 & 10.4000 & 0.9129 \\ \midrule
\texttt{Avg.} & - & - & 16 & 0.7884 & 0.8612 & 0.8193 & 0.7872 & 4.9429 & 11.54 & 0.8983 \\

\bottomrule
\end{tabular}
\end{table}

\section{Conclusion}

In this work, we investigated the utility and implications of employing large language models, particularly OpenAI’s ChatGPT, within the healthcare sector. We aimed to demonstrate their potential role in enhancing decision-making processes, drawing particular attention to the use of contextual prompts for high-quality predictions and the value of integrating domain-specific knowledge from interpretable Machine Learning (ML) models. 

Our analysis affirmed the strength and promise of OpenAI’s ChatGPT for clinical decision-making. In \textit{zero-shot scenarios}, its initial performance was found to lag behind classical ML models. However, with an increase in the number of examples used for prompt construction, i.e., in \textit{few-shot scenarios}, ChatGPT showcased the ability to improve significantly, reaching, and in some instances surpassing, the performance of traditional supervised ML models. This capacity to learn and adapt with additional examples emphasizes the potential of these models in contexts with limited data.

A key finding from our study was the notable performance improvement in ChatGPT when domain knowledge was integrated, specifically prediction outcomes from high-performing ML models such as \XGBoost. This underscores the value of harnessing domain knowledge information and corroborates our hypothesis that expert knowledge (here obtained through \XGBoost) provides the beneficial domain knowledge input. Such integration of AI with medical expertise holds immense potential for healthcare applications, illustrating the ability of AI models to leverage traditional ML insights.

However, we identified considerable variability in the performance of OpenAI models, along with the potential risk of higher False Positives (FP) even though False Negatives (FN) were generally lower compared to traditional ML models. n the medical field, both types of errors have serious implications,\textit{ however, the cost of FN can sometimes be particularly high}, such as in critical diagnoses like cancer, where a missed detection could lead to dire consequences. On the other hand, a higher rate of FP, as seen in the OpenAI models, while concerning, could be viewed as a safer error direction in these high-stakes situations. In general, both types of incorrect predictions bear significant implications within a medical context, introducing over-diagnosis, and unnecessary treatments, causing physical, emotional, and financial burdens to patients.

Looking ahead, future endeavors should persist in refining the design of prompts, mitigating social and ethical risks \cite{deldjoo2023fairnesschatgpt,deldjoo2023navigate,weidinger2021ethical}, and optimizing performance. An in-depth examination of zero-shot and few-shot learning dynamics would offer valuable insights for designing more reliable AI systems. Expanding the application of these techniques to different healthcare realms could broaden the impact of AI-assisted decision-making tools. Given the potential risks, we suggest the cautious use of these models in clinical settings, accentuating the importance of careful model design and implementation. Ultimately, the seamless blending of AI with domain-specific expertise will be the key to successfully deploying large language models within the healthcare sector.

\bibliographystyle{splncs04}
\bibliography{mybibliography}
\end{document}